# Compound Fréchet Inception Distance for Quality Assessment of GAN Created Images

Eric J. Nunn[1], Pejman Khadivi[1], Shadrokh Samavi[2]
[1]Department of Computer Science, Seattle University, Seattle, WA, USA
[2]Department of Elect. & Comp. Eng., McMaster University, Hamilton, ON, Canada

## Abstract

Generative adversarial networks or GANs are a type of generative modeling framework. GANs involve a pair of neural networks engaged in a competition in iteratively creating fake data, indistinguishable from the real data. One notable application of GANs is developing fake human faces, also known as "deep fakes," due to the deep learning algorithms at the core of the GAN framework. Measuring the quality of the generated images is inherently subjective but attempts to objectify quality using standardized metrics have been made. One example of objective metrics is the Fréchet Inception Distance (FID), which measures the difference between distributions of feature vectors for two separate datasets of images. There are situations that images with low perceptual qualities are not assigned appropriate FID scores. We propose to improve the robustness of the evaluation process by integrating lower-level features to cover a wider array of visual defects. Our proposed method integrates three levels of feature abstractions to evaluate the quality of generated images. Experimental evaluations show better performance of the proposed method for distorted images.

## 1-Introduction

Generative adversarial networks or GANs are a framework of generative modeling. GANs consist of two multilayer perceptrons that compete against each other in an adversarial manner to generate novel data indistinguishable from real data. GANs were first proposed by Ian Goodfellow et al. [1] in 2014.

GANs consist of two neural networks defined as a "generator" and a "discriminator." The generator's distribution $p_x$ over data $x$ is learned by providing the generator with an input noise vector $z$ to generate a prior $p_z(z)$. The prior is then mapped to data space by the function $G(z;\theta_g)$, where $G$ is a differentiable function represented by a multilayer perceptron with parameters $\theta$. The discriminator is defined by a separate multilayer perceptron $D(x;\theta_d)$ which outputs a single scalar. The output of $D$ represents the probability that $x$ came from the real data rather than $p_g$.

The discriminator $D$ is trained to maximize the probability of correctly assigning the label to both training examples and generated samples from $G$. The generator is simultaneously trained to

minimize the *log(1 – D(G(z))*. The two networks compete in a minimax game with a value function defined as:

$$\min_G \max_D V(D, G) = \mathbb{E}_{x \sim p_{data}(x)}[\log D(x)] + \mathbb{E}_{z \sim p_z(z)}[1 - \log D(G(x))]. \quad (1)$$

Training is finished when the resulting generator distribution $p_z$ matches the distribution of the training data. When deployed into production, the discriminator is discarded, and only the generator is used to create novel data.

Applications for GANs range from image-to-image translation [6][7] to super-resolution [8], but their most well-known application relates to "deepfakes." The term deepfake is a combination of the phrases "deep learning" and "fake," which refers to the alteration of media that appears 1authentic to the human eye. The most common form of deepfake involves images of humans. These deepfake images represent one of the most challenging applications for GANs since human evaluators have an extremely acute ability to distinguish between real and fake faces. GANs trained to generate human faces, therefore, are burdened with increased scrutiny when evaluated.

One issue inherent to generative models such as GANs is the subjective nature of evaluating their output quality. Quantitative measures of image quality are difficult to define, which causes issues when determining when a model is ready to deploy into production. Many attempts have been made to define objective metrics for evaluating GAN quality, but current metrics fall short in specific situations when compared to the subjective evaluation by humans. The following section describes several methods for assessing GANs using quantitative metrics.

## 2-Related Work

One of the most widely adopted metrics, proposed by Salimans et al. [9], is the Inception Score (IS), which evaluates GANs by using a pre-trained network (InceptionNet [10], trained on the ImageNet [11] dataset) to capture the desired properties in generated images. It measures the KL divergence between the conditional label distribution *p(y|x)* and marginal label distribution *p(y)* obtained from a sample of generated images. Low entropy for *p(y|x)* suggests higher image quality and easily classifiable samples. In contrast, high entropy for *p(x)* suggests high diversity, assuming all classes are equally represented in the set of samples.

$$\exp(\mathbb{E}_x[\mathbb{KL}(p(y|x) \| p(y))]) = \exp(H(y) - \mathbb{E}_x[H(y|x)]), \quad (2)$$

where *p(y|x)* is the conditional label distribution for image *x* estimated using a pre-trained Inception model [10], and *p(y)* is the marginal distribution: $p(y) \approx \frac{1}{N}\sum_{n=1}^{N} p(y|x_n = G(z_n))$. $H(x)$ represents entropy of variable *x*.

Ali Borji [3] describes several limitations of the Inception Score (IS). It tends to favor a "memory GAN" that stores all training samples, thus is unable to detect overfitting. This issue is worsened by the fact that IS does not make use of a holdout validation set. The inception score also ignores $P_r$ and only considers $P_g$. As a result, IS may favor models that simply learn sharp and diversified images instead of learning the true distribution $P_r$.

Fréchet Inception Distance (FID) [12] takes a step further than the inception score and is one of the most common metrics used to evaluate GANs today. FID embeds a set of sample images into a feature space representing a high-level abstraction. When viewed as a continuous multivariate Gaussian distribution, this feature space is used to compute the mean and variance of the generated data and real data. The Fréchet distance between these two distributions is used to evaluate the quality of generated samples, where a lower FID means a smaller distance between real and generated distributions. FID is calculated using the following equation:

$$FID(r, g) = \left\| \mu_r - \mu_g \right\|_2^2 + Tr(\Sigma_r + \Sigma_g - 2\sqrt{\Sigma_r \Sigma_g}), \qquad (3)$$

where ($\mu_r$, $\Sigma_r$) and ($\mu_g$, $\Sigma_g$) are the mean and covariance of the real data and model distributions, respectively.

As with the Inception Score, FID obtains the feature vector from an intermediate layer of Inception-V3 [10]. InceptionNet is an image classifier that employs an advanced type of neural network architecture called an inception block or inception module. This module combines multiple, parallel convolution layers with differing kernel sizes to better learn complex features.

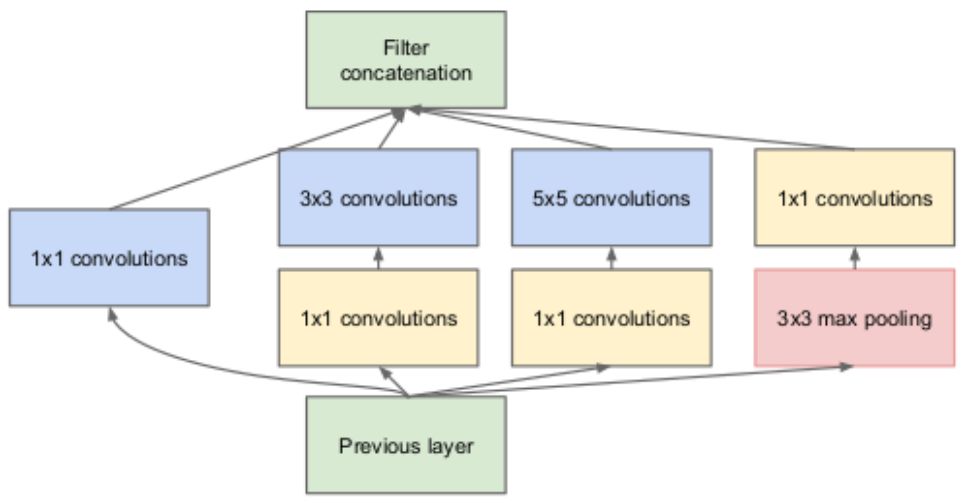

*Figure 1: Inception module architecture. Convolutions with various kernel sizes allow the internal layers to learn which kernel size is best suited to identify important image features.*

At the end of each inception block is a pooling layer to reduce the dimensionality of the feature maps. The final pooling layer of Inception-V3, directly before the final classification layer, is used as the feature vector for the FID metric. This layer is used because it represents the highest level of abstraction made by the model and contains global information relating closest to the objects present in the image.

FID is an improvement over IS because it can detect intra-class mode dropping, *i.e., a model that generates only one image per class can score a high IS but will have a bad FID* [9]; however, FID has its limitations. It is biased towards global features as opposed to local features. This is due to its use of high-level feature abstractions taken from Inception-V3, so lower-level features such as texture are ignored.

## 3-Compound Fréchet Inception Distance

This section presents an algorithm that expands upon the original FID metric [12] by combining FID scores from three feature vectors taken from the three inception blocks in the Inception-V3 [12]. We call our method "Compound Fréchet Inception Distance" or CFID. The three unique feature vectors give us a broad spectrum of information from multiple levels of abstraction, giving an insight into both local and global features of an image.

Local features are used in low-level image recognition tasks [15] and usually consist of local gradients. We also consider mid-level features such as lines and edges. In contrast, global features are typically used for object detection and classification tasks [15] and consist of object-level features. Information from low-level to high-level features would provide a more robust view of the image at the expense of computational overhead.

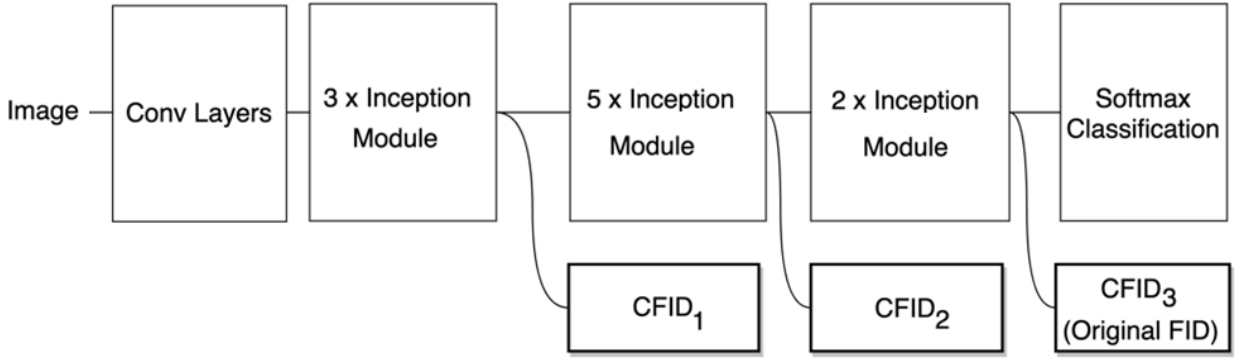

*Figure 2: Inception-V3 block diagram showing all three CFID feature vectors extracted from the pooling layers at the end of each inception module. CFID numbers refer to the order of their corresponding inception block.*

The feature vectors taken from Inception-V3 include max-pooling layer 1 (MaxPool1), max-pooling layer 2 (MaxPool2), and average-pooling layer (AvgPool). Layer names for Inception-V3 are indicated in [13]. Each of these layers is located at the end of each inception block in the network. Figure 2 shows a simplified block diagram showing the location of each of these layers. The naming convention for each CFID score refers to the order of their corresponding inception block within Inception-V3. $CFID_3$ is therefore identical to the original FID in [12]. $CFID_1$ and $CFID_2$ represent the new FID scores based on low and medium-level features. Equation (4) defines the calculation for each CFID score:

$$CFID_i(r, g) = \|\mu_{ri} - \mu_{gi}\|_2^2 + Tr\left(\Sigma_{ri} + \Sigma_{gi} - 2\sqrt{\Sigma_{ri}\Sigma_{gi}}\right), \quad (4)$$

for $i \in \{1,2,3\}$, and ($\mu_r$, $\Sigma_r$) and ($\mu_g$, $\Sigma_g$) are the mean and covariance of the real data and model distributions, respectively.

Each of the feature vectors taken from Inception-V3 has a different shape, and therefore their CFID statistics must be normalized to allow an equal weighting for each abstraction level. We want to keep CFID comparable to the original FID metric. Therefore, the three separate FID scores are normalized by scaling each score to match the original FID [12] feature vector's shape of (2048, 1, 1).

Table I: Inception-V3 layer properties

| Layer Name | Layer Shape | Total Features |
|---|---|---|
| MaxPool1 ($CFID_1$) | (64, 73, 73) | 341,056 |
| MaxPool2 ($CFID_2$) | (192, 35, 35) | 235,200 |
| AvgPool ($CFID_3$) | (2048, 1, 1) | 2,048 |

As seen in Table 1, MaxPool1 and MaxPool2 are 3-dimensional tensors. On the other hand, AvgPool is a 1-D vector. The tensors for $CFID_1$ and $CFID_2$ are flattened to create a 1-D vector to calculate each layer's FID. The mean and covariances ($\mu_r$, $\Sigma_r$) and ($\mu_g$, $\Sigma_g$) are computed for each feature vector. Finally, the covariances are scaled using the ratio of features between MaxPool1 or MaxPool2 and AvgPool for $CFID_1$ and $CFID_2$, respectively. The three CFID values are kept separate and distinct at this time to compare each value and determine their responses to changes in the underlying image distributions.

## 4-Experiments and Analysis

*A. Preparation*
Our experiments were structured to reflect the original experiments conducted by M. Heusel et al. [11] in their preliminary work on FID. We show how CFID compares to the original FID metric when various local and global distortions are applied to an image **X**. The distortions we applied are as follows:

1. **Gaussian noise**: We constructed a matrix **N** with Gaussian noise scaled from [0,255]. The noisy image was computed as $(1 - \alpha)X + \alpha N$ for $\alpha \in \{0, 0.05, 0.1, 0.2\}$. The larger $\alpha$ is, the more noise is added to the image.
2. **Gaussian blur**: We convolved an image with a Gaussian kernel with standard deviation $\alpha \in \{0, 1, 2, 4\}$. The larger $\alpha$ is, the more the image is blurred.
3. **Spiral warp**: The image is distorted by warping the center with a spiral (whirlpool effect). Consider the coordinate *(x, y)*; warping is performed by computing the reverse mapping for the swirl transformation, which gives the location which is mapped to *(x, y)*. The polar coordinates are computed relative to a center *(x0, y0)* given by the angle $\theta = \arctan(\frac{y-y_0}{x-x_0})$ and the radius $r = \sqrt{(x-x_0)^2 + (y-y_0)^2}$. The coordinates are transformed according to $\theta' = \theta + \alpha e^{-5r/(\ln 2\rho)}$. The parameter $\alpha$ is a parameter for the amount of swirl, and $\rho$ is a parameter for the radius of swirl measured in pixels. The original coordinates, where the color for *(x, y)* can be found, are $x_{org} = x_0 + r\cos(\theta')$ and $y_{org} = y_0 + r\sin(\theta')$. We set *(x0, y0)* to the center of the image and $\rho = 25$. The distortion level is given by the amount of swirl $\alpha \in \{0, 1, 2, 4\}$. The larger $\alpha$ is, the larger the amount of swirl in the image.
4. **Salt and pepper noise**: Individual channels of some pixels in the image are set to black or white with a 50% probability for each. The probability of each pixel channel being chosen is given by the noise parameter $\alpha \in \{0, 0.1, 0.2, 0.3\}$. The larger $\alpha$ is, the higher is the probability of a pixel becoming 0 or 255.

*B. Experimental Results*
**Evaluation on a single image.** Each pack of the CFID values was calculated for two sets of images. Some sample results of our experiments are shown in the following figures. CFID components are kept separate to show individual responses to image distortion.

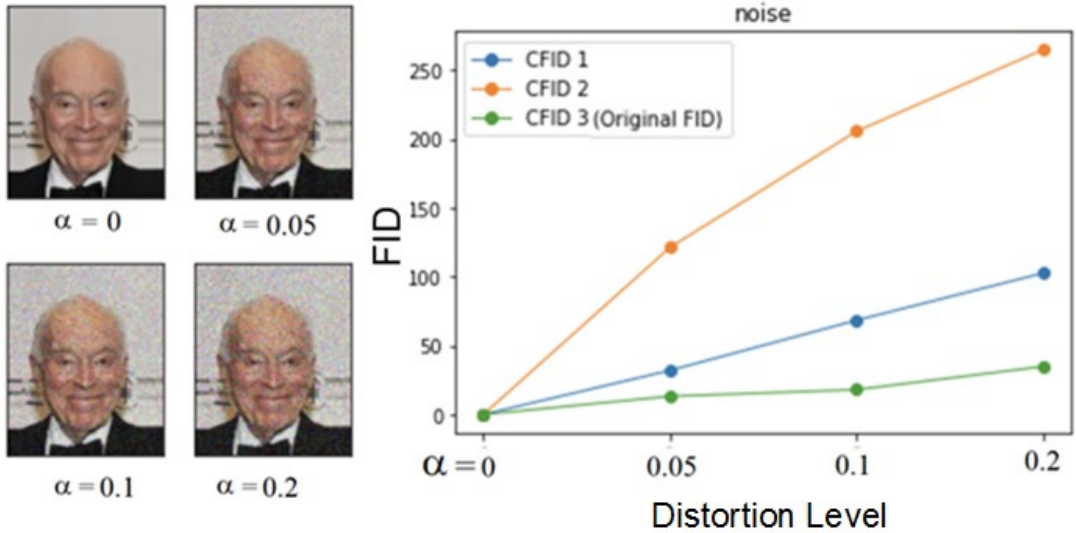

*Figure 3: CFID results for an image with increasing levels of Gaussian noise.*

In Fig. 3, we see that $CFID_1$ and $CFID_2$ follow a steady increase in the calculated distance. The distance increase is consistent with the increasing Gaussian noise. In contrast, we do not see a meaningful increase in the original FID scores ($CFID_3$).

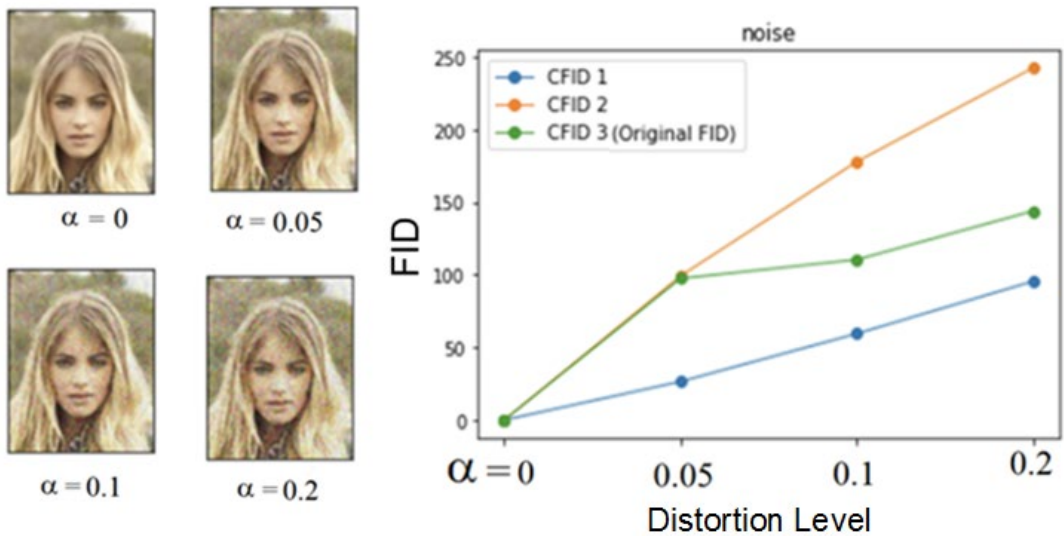

*Figure 4: CFID results for an image with increasing levels of Gaussian noise.*

Figure 4 shows another example of an image that is distorted with increasing Gaussian noise. We see that $CFID_2$ shows a meaningful representation of falling image quality while the original FID gets saturated when the level of distortion passes α=0.1. The Gaussian noise has distinguishable effects on the mid-level image features and can be detected by the proposed $CFID_2$.

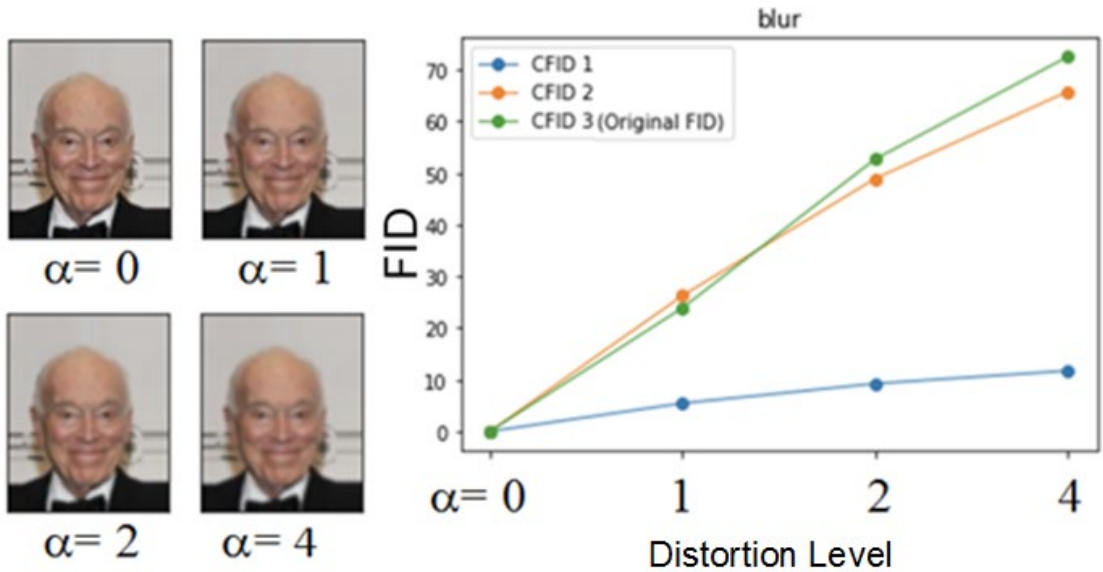

Figure 5: *An example for CFID results for increasing levels of Gaussian blur.*

Figure 5 shows the effect of the Gaussian blur distortions on different CFID measures. It can be seen that Gaussian blur is a high-level distortion, and global aspects of the image are affected. Hence, for this type of high-level distortion, we should use high-level features represented by $CFID_3$. In contrast, we see that $CFID_1$ has low sensitivity to high-level disturbance.

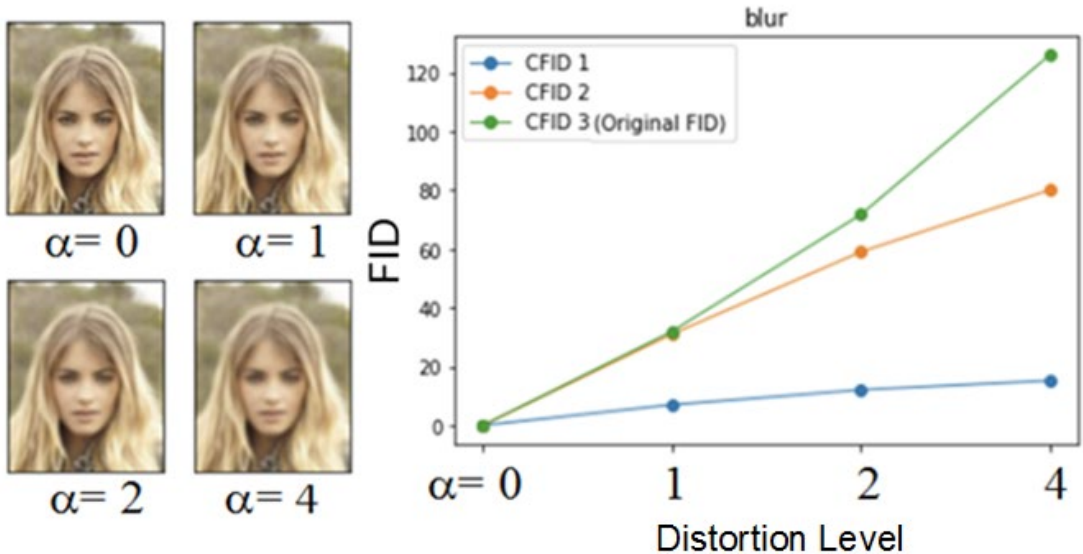

Figure 6: *An example for CFID results for increasing levels of Gaussian blur.*

In Figure 6, another example shows the effects of global distortions of the Gaussian blur. Unlike Fig. 5, which has a smooth background, variance is more affected by the noise in this image. Hence $CFID_3$ is more distinctive to show the decay in the quality of the image.

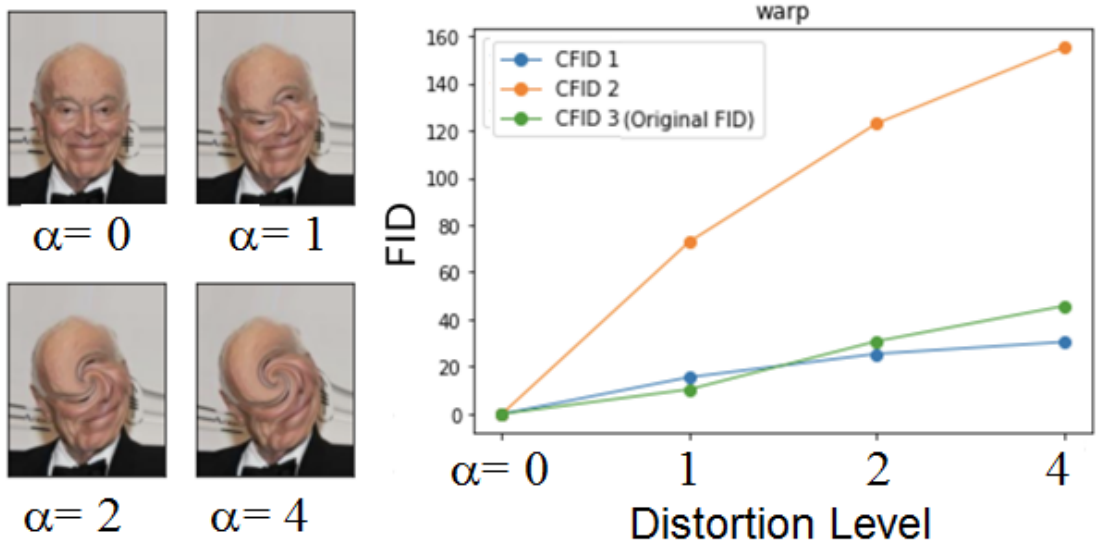

*Figure 7: CFID results for increasing levels of spiral warping.*

Figure 7 has a smooth background with a small variance. Hence, global variance is not affected by the warping distortion, and $CFID_3$ is not showing high sensitivity to such changes. In contrast, we see that mid-level features that $CFID_2$ represents, show a more distinctive measure of distortions.

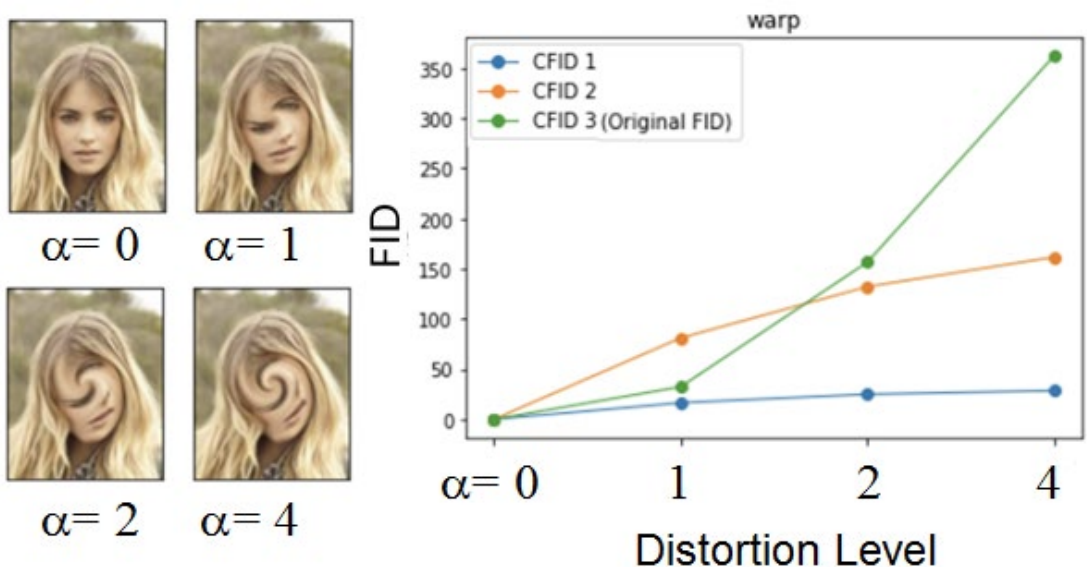

*Figure 8: CFID results for increasing levels of spiral warping.*

Figure 8 is an example where all parts of the image have high variance. In such images, warping distortions are not affecting low and mid-level features. The amount of warping is well represented by global features that are extracted by the $CFID_3$ criterion.

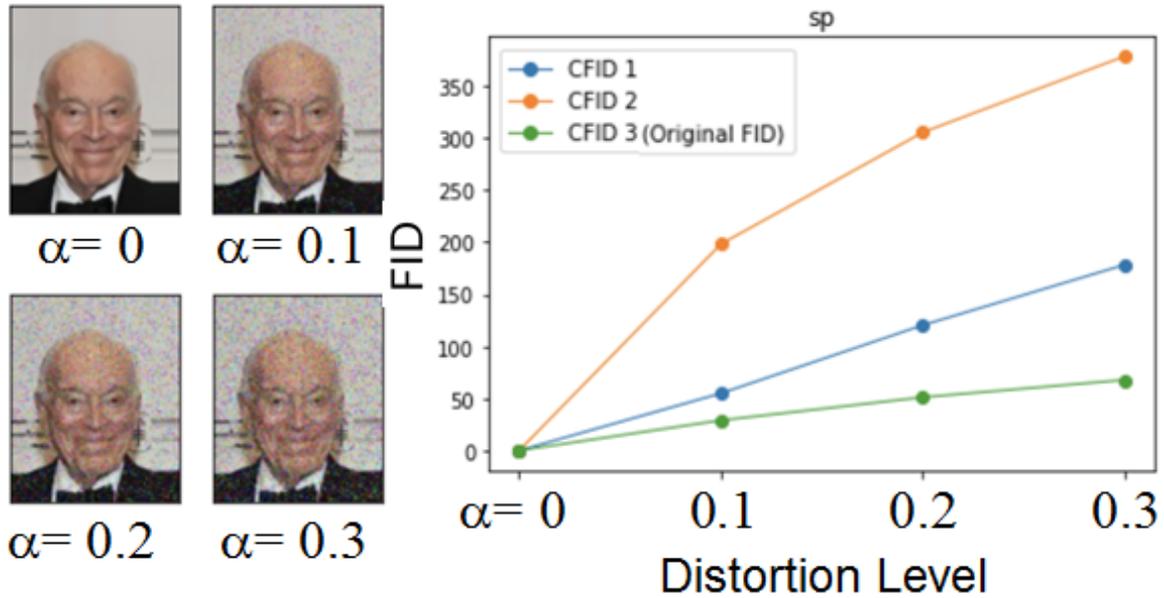

*Figure 9: CFID results for increasing levels of Salt & Pepper Noise.*

Salt and pepper noise is a local disturbance and can be extracted by low and mid-level features. In Fig. 9, we see that $CFID_1$ and $CFID_2$ are showing meaningful distances. Hence, for the salt and pepper noise, we can use $CFID_2$.

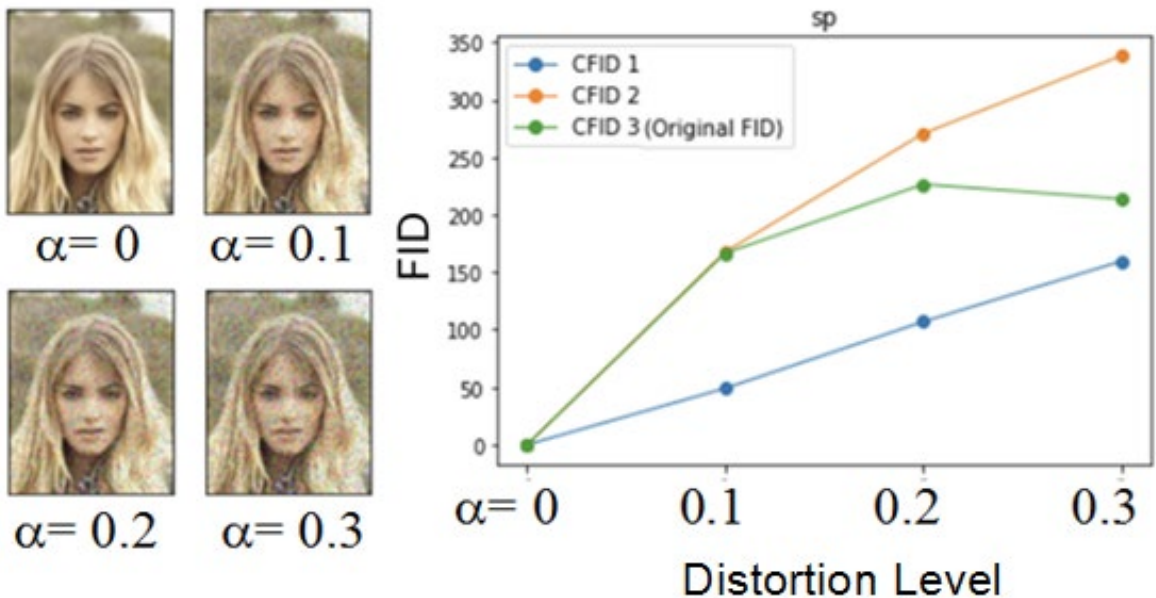

*Figure 10: CFID results for increasing levels of Salt & Pepper Noise.*

Figure 10 shows that global features are affected by local distortions. But as the level of distortion increases, the global changes of the mean and variance get saturated. Hence, the original FID is not appropriate to show the level of disturbance.

# 5-Conclusion

To improve quantitative GAN evaluation, we have introduced the "Compound Fréchet Inception Distance" (CFID). The proposed CFID captures the similarity of generated images to real ones better than the original FID because it compares the image distributions based on multiple levels of feature abstraction. The initial experimental results show that each CFID score responds differently to each type of image distortion. Our experiments show that different layers of Inception-V3 respond differently based on the kind of distortions. Having a set of three CFID scores gives us a broader perspective to GAN image quality by analyzing multiple levels of feature extraction, and the combination of these three CFID scores correlates more with typical human evaluation.

Through experiments, we showed that the effects of distortion depend on three factors. First, the amount of distortion should be considered. Second, the type of distortion is important. Some distortions have local effects on the distribution of information within an image, while some image distortions have global consequences. Third, the distribution of the pixels plays a role in how the distortion affects the image. We showed that extracting low, medium, and high-level features is the solution in measuring a robust distance between a distorted and normal image.

Larger-scale experiments are needed to say which CFID score is best for different applications definitively. However, our preliminary results show that the current FID metric is not sufficient to capture all types of image distortions and does not always match human interpretability.

As a result, we propose the following function for the image quality assessment.

$$CFID = \max(CFID_i(r, g)) \quad (5)$$

We see that by choosing the maximum of the three CFIDs we will get the most appropriate distance between the actual and distorted image. The appropriateness of the span is irrespective of the distortion and the image distribution type.

# 6-References